\documentclass{article} 
\pdfoutput=1
\usepackage{iclr2020_conference,times}

\usepackage{amsmath,amsfonts,bm}

\def\eqref#1{equation~\ref{#1}}

\def\1{\bm{1}}

\DeclareMathAlphabet{\mathsfit}{\encodingdefault}{\sfdefault}{m}{sl}
\SetMathAlphabet{\mathsfit}{bold}{\encodingdefault}{\sfdefault}{bx}{n}

\usepackage{hyperref}
\usepackage{url}
\usepackage{graphicx}
\usepackage{subfigure}

\title{Image-based phenotyping of diverse Rice \\ (\textit{Oryza Sativa} L.) Genotypes}

\author{Mukesh Kumar Vishal\textsuperscript{1 \textdagger}, 
Dipesh Tamboli\textsuperscript{1},
Abhijeet Patil\textsuperscript{1}, 
Rohit Saluja\textsuperscript{1}, 
\\ 
\textbf{Biplab Banerjee\textsuperscript{1 \textdaggerdbl}, 
Amit Sethi\textsuperscript{1}, 
Dhandapani Raju\textsuperscript{2}, 
Sudhir Kumar\textsuperscript{2},
R N Sahoo\textsuperscript{2}},
\\
\textbf{Viswanathan Chinnusamy\textsuperscript{2}, and 
J Adinarayana\textsuperscript{1 \textdollar}} \\  
\textsuperscript{1} Indian Institute of Technology Bombay, India \\
\textsuperscript{2} ICAR-Indian Agricultural Research Institute 
New Delhi, India \\
\textsuperscript{\textdagger}\texttt{174314001@iitb.ac.in}, \textsuperscript{\textdaggerdbl}\texttt{getbiplab@gmail.com},
\textsuperscript{\textdollar}\texttt{jadi.iitb@gmail.com}}

\iclrfinalcopy 
\begin{document}
\maketitle
\begin{abstract}
Development of either drought-resistant or drought-tolerant varieties in rice (\textit{Oryza sativa } L.), especially for high yield in the context of climate change, is a crucial task across the world. The need for high yielding rice varieties is a prime concern for developing nations like India, China, and other Asian-African countries where rice is a primary staple food. The present investigation is carried out for discriminating drought tolerant, and susceptible genotypes. A total of 150 genotypes were grown under controlled conditions to evaluate at High Throughput Plant Phenomics facility,  Nanaji Deshmukh  Plant Phenomics  Centre,  Indian  Council of  Agricultural  Research-Indian  Agricultural  Research  Institute, New Delhi. A subset of 10 genotypes is taken out of 150 for the current investigation. To discriminate against the genotypes, we considered features such as the number of leaves per plant, the convex hull and convex hull area of a plant – convex hull formed by joining the tips of the leaves, the number of leaves per unit convex hull of a plant, canopy spread - vertical spread, and horizontal spread of a plant. We trained You Only Look Once (YOLO) deep learning algorithm for leaves tips detection and to estimate the number of leaves in a rice plant. With this proposed framework, we screened the genotypes based on selected traits. These genotypes were further grouped among different groupings of drought-tolerant and drought susceptible genotypes using the Ward method of clustering.
\end{abstract}
\section{Introduction}
Rice (Oryza sativa L.)  is a primary staple food in the world \cite{tubiello2013faostat}. China and India are the leading rice-producing nation in the world. These two countries with other Asian nations contribute 90\% of world rice production \cite{muthayya2014overview}. In current climate changing scenarios, food security is the prime concern in these highly populated developing nations, especially India and China \cite{tubiello2013faostat}, \cite{muthayya2014overview}.  To cater to the need of the rising population in these countries, constant efforts are being made to achieve high yield through genetic improvements programs. To mitigate the issues arising due to change in the climate in recent decades (IPPC, 2007, \cite{cruz2007asia}) demands climate-resilient genotype for these crops. So there is a need to identify the donor rice genotypes with higher precision in breeding programs. Conventionally, in traditional phenotyping, the often-prevalent recurrent selection breeding techniques are applied for the initial screening and evaluation of genotypes to develop a new variety  \cite{dhondt2013cell}. Moreover, traditional phenotyping involves mostly human senses, which limits high efficiency, and are prone to human error. These approaches are destructive, invasive, time consuming, less effective and restricts to achieve high precision. Furthermore, for high precision observations and combating these bottlenecks of traditional phenotyping, the High Throughput Plant Phenotyping evolved and has many advantages \cite{cobb2013next}, \cite{seelig2008assessment}, \cite{araus2014field}. The HTPP applies various sensors, imaging techniques, and Remote Sensing (RS) principles for recording plant observations and weather parameters (environmental variability) for phenotyping in an automatic and semi-automatic manner \cite{araus2014field}, \cite{bai2016multi}, \cite{ghanem2015physiological}. Many morphological traits like plant height, plant canopy spreading type, convex hull, plant leaf area, number of leaves in a plant, color, etc. are used for phenotyping \cite{bardenas1965morphology}, \cite{upov1987guidelines}. These traits are also used for grouping of the genotypes and understanding the diversity among genotypes as well as species. 

In this study, we have considered many image-based traits like the number of leaves per plant, the ratio of the number of leaves per plant to the per unit area of its convex hull, area of the convex hull for each plant, the plant spread (canopy spread) based on the top view of the plant images, to classify the diverse rice (Oryza sativa L.) genotypes based on the top view of the plant. For leaf counting, we used deep learning techniques, You Only Look Once (YOLO) \cite{redmon2013darknet}. YOLO is convolutional neural networks based state-of-the-art object detection algorithm \cite{redmon2017yolo9000}, \cite{redmon2018yolov3}, \cite{ren2015faster}. We employed YOLO for leaf detection owing to its speed for real-time object detection. Apart from this, the convex hull of all genotype, vertical spread, and horizontal spread of the canopy is calculated for diversity analysis.
\begin{figure}[ht] 
\centering
  \subfigure[Rice genotypes in a controlled greenhouse]{%
    \includegraphics[width=6cm]{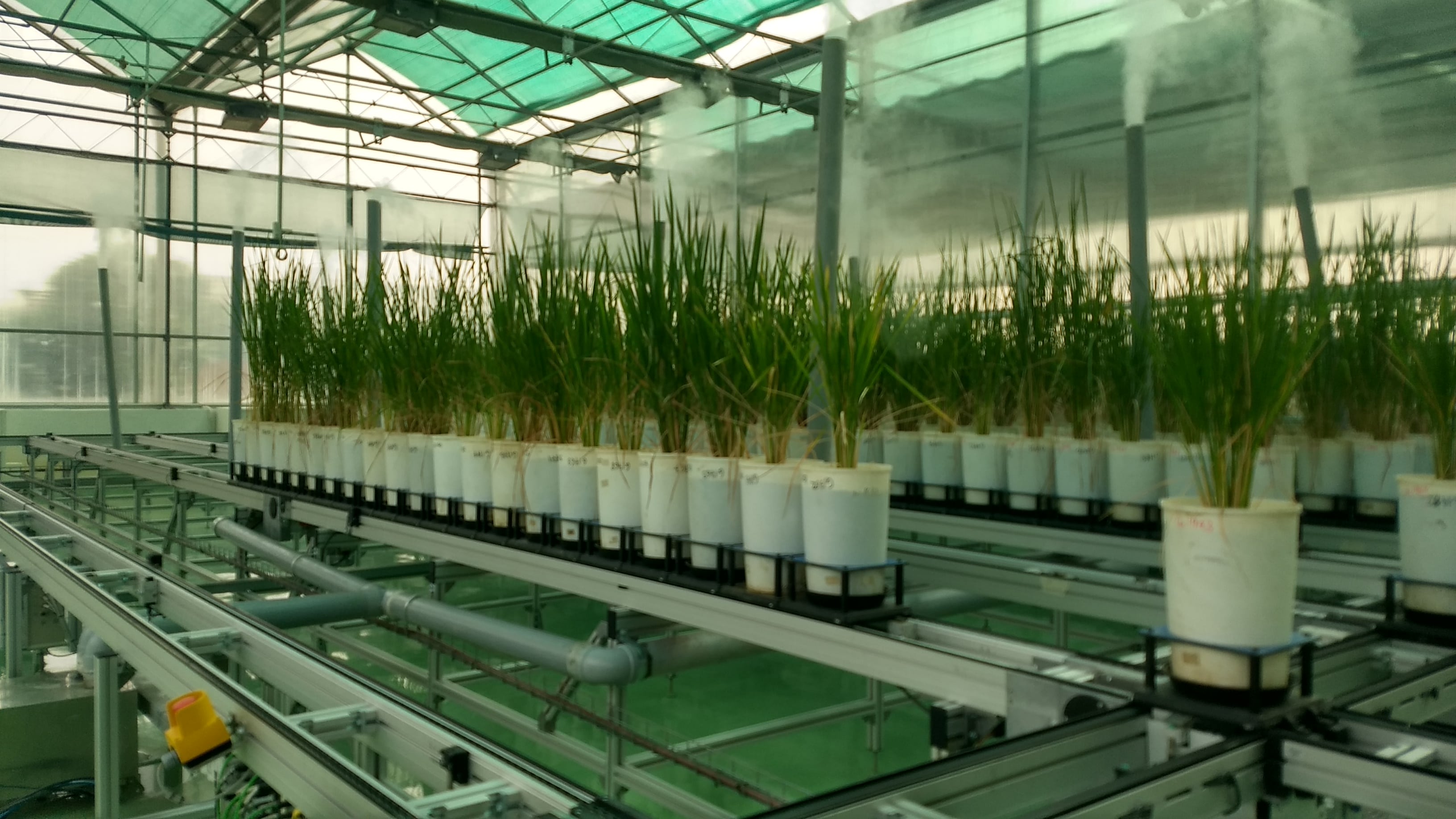} 
  } 
  \quad
  \subfigure[Rice plants on the car, moving for imaging]{%
    \includegraphics[width=6cm] {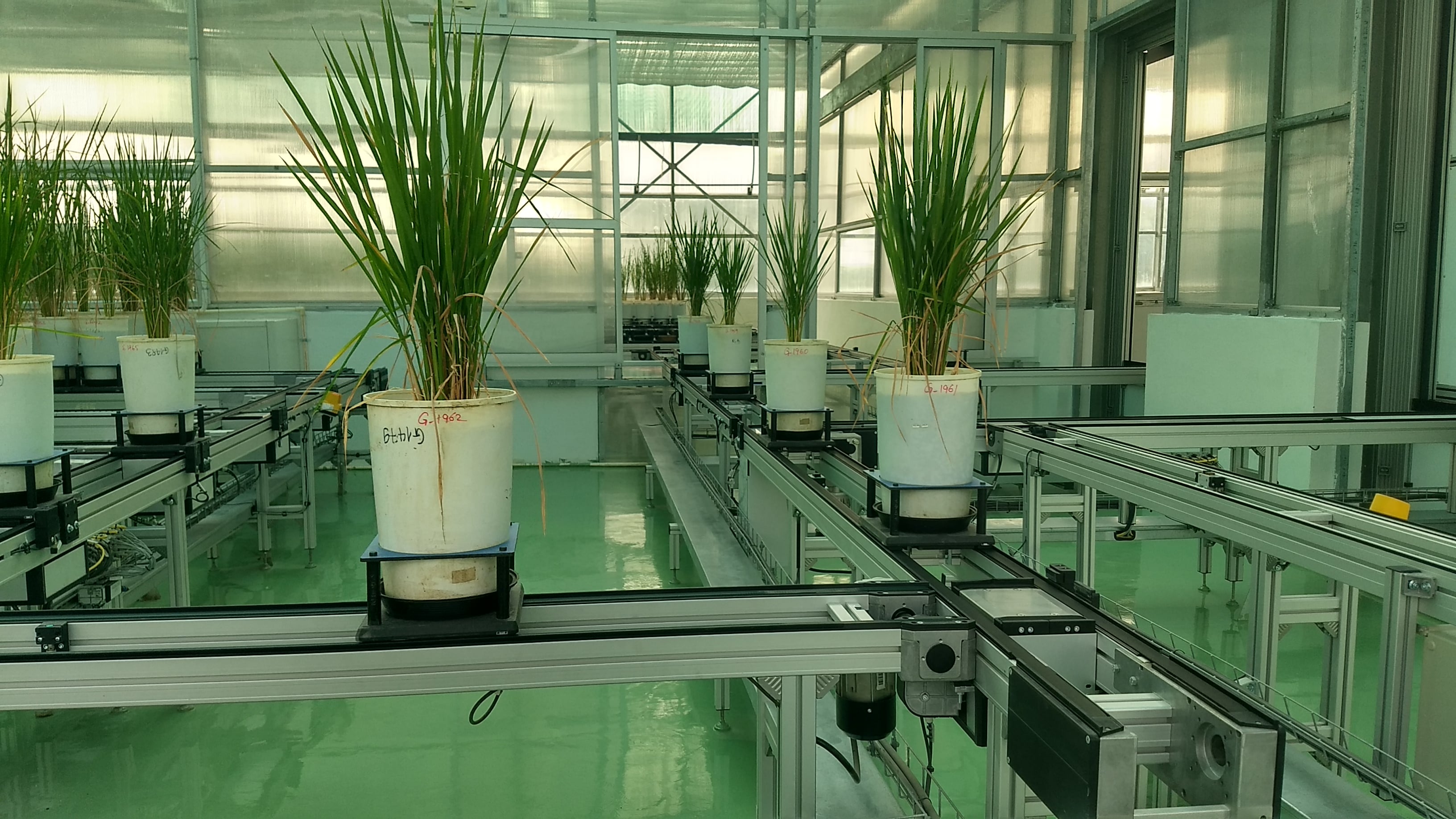}
  }
  \caption{Experimental set up  at the Nanaji Deshmukh Plant Phenomics Centre (NDPPC)} 
  \label{fig:setup}
\end{figure}   
\section{Dataset and Methods}
\subsection{Dataset}
The experimental setup is shown in Figure \ref{fig:setup}. In this study, datasets are generated for diverse rice genotypes grown in pots in the climate-controlled greenhouses at Nanaji Deshmukh Plant Phenomics Centre (NDPPC), Indian Council of Agricultural Research-Indian Agricultural Research Institute (ICAR-IARI) located in New Delhi, India. The data is a subset of 150 rice genotypes grown for Genome-Wide-Association-Study (GWAS) to identify the major Quantitative Trait Loci (QTL) responsible for drought tolerance. Pot culture experiment was conducted with one set of plants (3 replications for each genotype) with moisture-deficit stress (~12.5 \% soil moisture content) during maximum tillering stage (35 to 55 days after transplanting), while the other set grown in well watered (25\%) condition. The RGB visual images of pot-grown rice plants of each selected genotypes were captured using a visual camera (Prosilica GT6600 series for monochrome and color image,RGB of spectral range 400-700 nm)installed in scanalyzer3D, LemnaTec, imaging platform (LemnaTec GmbH, Aachen, Germany) at NDPPC. We capture the top view and side view images of plants having a size $6576 \times 4384$ recorded over an interval of five days. The rice genotypes were selected for high polymorphic variation in the leaf number.
\begin{figure}[ht] 
\centering
  \subfigure[RGB image of a rice plant, top view]{%
    \includegraphics[width=4cm]{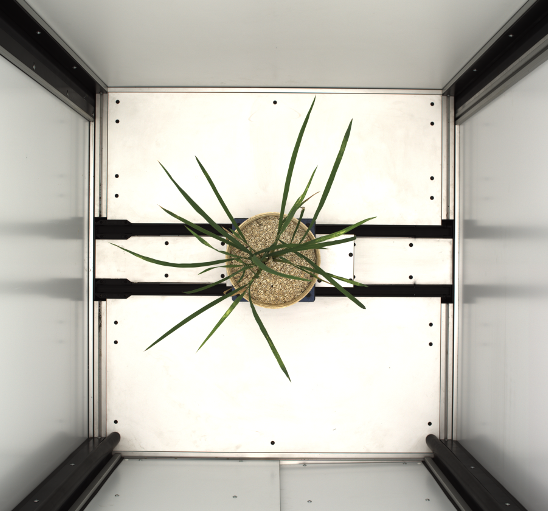} 
  } 
  \quad
  \subfigure[Annotations of leaf tip for a rice plant]{%
    \includegraphics[width=4cm]{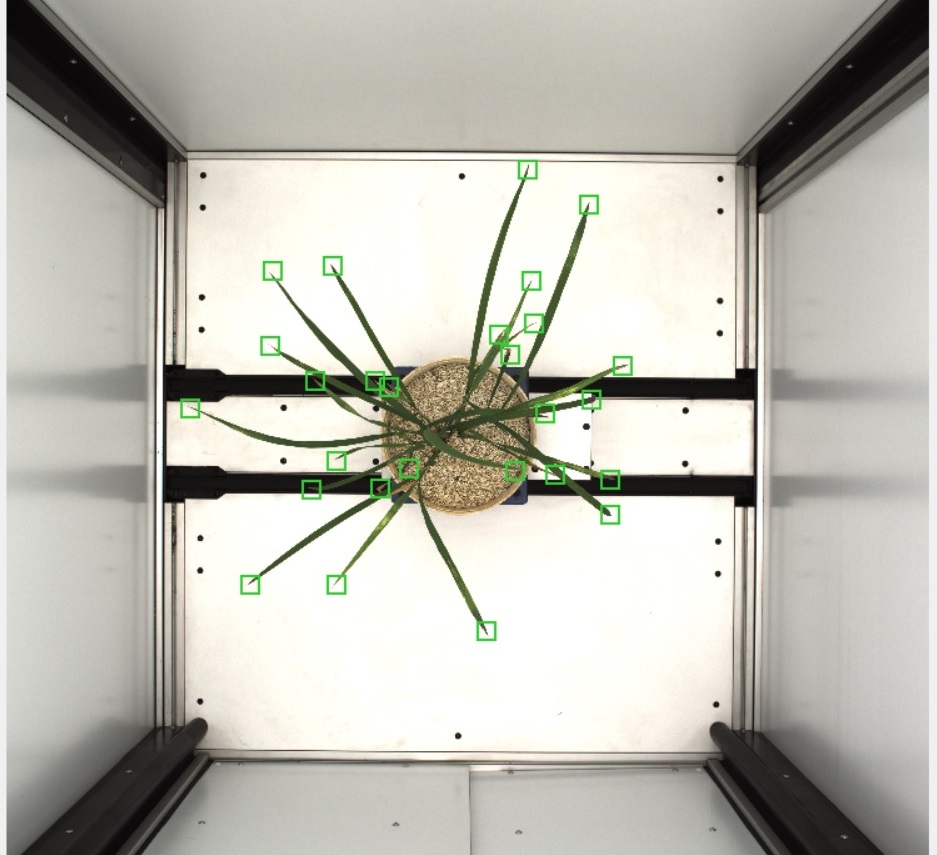}
  }
  \caption{Data annotation procedure} 
  \label{fig:annotation}
\end{figure} 
\subsection{Methods}
\begin{figure}[t]
\centering
  \subfigure[]{%
    \includegraphics[width=6.5cm]{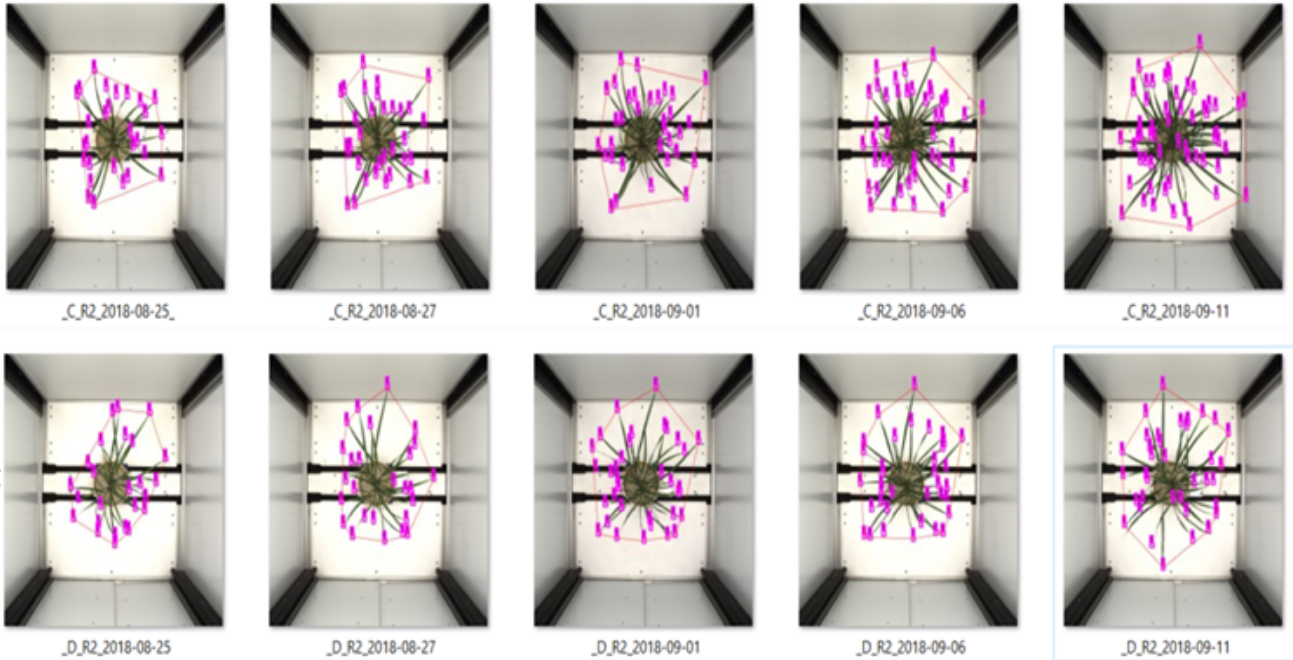} 
  } 
  \quad
  \subfigure[]{%
    \includegraphics[width=6.5cm]{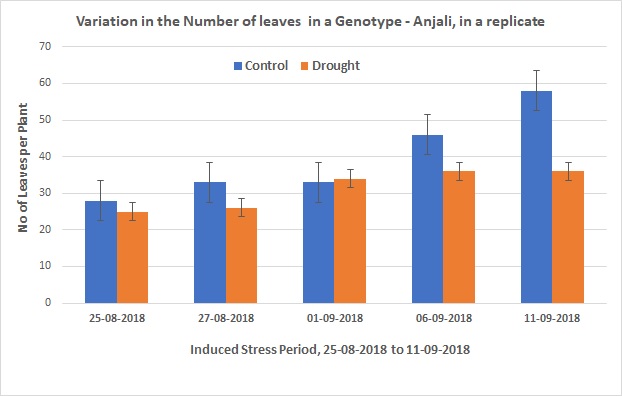}
  }
  \caption{(a) Variations in the number of leaves, and the number of leaves per unit area of the convex hull of a plant for a genotype in control (top) and drought (bottom) condition. (b) Figure shows variation in the number of leaves for Anjali genotype in control and drought conditions during stress period (35 to 55 DAT)} 
  \label{fig:results_1_samples}
\end{figure}

The original images were pre-processed and resized for the study.  After pre-processing, around 300 images were annotated, as shown in Figure \ref{fig:annotation}. These images are used for the training deep learning model, YOLO \cite{redmon2013darknet}, \cite{redmon2017yolo9000}, \cite{redmon2018yolov3}. YOLO has been implemented for leaves tip detection and estimation of the number of leaves. In the case of a rice plant, a leaf has a single tip. Therefore, a hypothesis was made that the number of leaves is equal to the number of tips for a rice plant. Few leaf counting works were done using convolution networks and de-convolution networks \cite{dobrescu2017leveraging}, \cite{paul2017count}, \cite{segui2015learning}, \cite{aich2017leaf}. Rice plant has a diverse plant architecture. Variation from sparse to dense leaves (canopy spread), and overlapping leaves create challenges for detection and segmentation of leaves in a rice plant. However, tips are rarely overlapped and are easily detectable, except leaves and their tips from few unproductive tillers that are crowded and hidden near culms and leaf sheath area in a rice plant just above-ground. The convex hull of each plant is calculated from predicted bounding boxes and their coordinates of leaves tips. This methodology has the advantage of drawing a convex hull simultaneously with predicted bounding boxes and their coordinates for train as well as test data. Few samples of leaves tips predicted by YOLO and convex hull formed by corresponding leaves tips are shown in Figure \ref{fig:results_1_samples}. The ratio of the number of leaves in a plant to the area of convex hull formed by leaf tips is considered to analyze plant architecture and understand canopy spreading architecture of rice plants under drought stress. When a plant goes under stress, then leaf rolling occurs and thus reduces the convex hull and compactness of the plant for spread type plants. Based on the number of leaves for each genotype and the number of leaves per unit area of the convex hull, we clustered the genotypes. This will help the breeder to infer about the phenotypic behavior of the genotype and assist in selecting the donors for further breeding programme. The clustering of genotype based on the number of leaves per unit area of the convex hull also has the potential for drought and agronomic study as we observe the vigour and canopy spread of the plant.
\section{Results and Discussion}
\begin{figure}[ht] 
\centering
  \subfigure[Considering all parameters]{%
    \includegraphics[width=6cm]{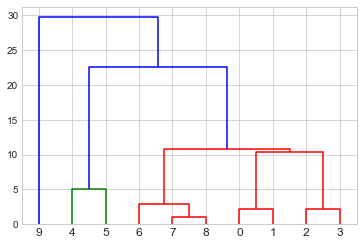} 
    \label{dendo1}
  } 
  \quad
  \subfigure[Considering the number of leaves per plant]{%
    \includegraphics[width=6cm]{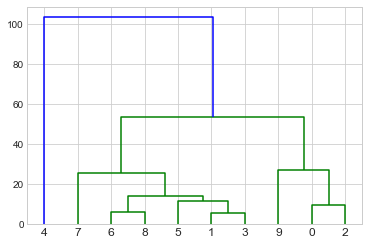}
    \label{dendo2}
  }
  \caption{Dendogram constructed with Ward clustering method for ten (10) genotypes of rice to discriminate drought and non-drought genotypes (0-ANJALI, 1-BLACKGORA, 2-ABHAYA X DAGADESI, 3-HEERA, 4-NAGINA 22, 5-RASI, 6-DULAR, 7-PMK-2, 8-SERATOES, 9-KALINGA-1)} 
  \label{fig:dendo}
\end{figure}  
A subset of 10 genotypes out of 150 genotypes was taken for current work, which has a total of approximately 400 images of these genotypes. After pre-processing the acquired images, we trained the YOLO for leaves tips detection. Leaf tip annotation, as shown in Figure 2. (b) for a rice plant as tip (object) detection, annotated tips are inside bounding boxes. These same predicted tips of each plants are used for calculation of convex hull for each plant for each replication of  genotypes i.e. plant in the experiment. We propose the number of leaves per unit area of the convex hull as a 2D trait in plant phenotyping. Figure \ref{fig:results_1_samples} depicts the variation of the number of leaves and convex hull for a genotype, Anjali, in controlled and drought conditions. We observed that the number of leaves setting in stress condition plants i.e. drought susceptible genotypes is less as compared to controlled plants as well as the drought tolerant plants. Also, for spread type plants, the convex hull was reduced in stress conditions. We observe that the number of leaves per unit area of the convex hull will be a potential trait to understand plant architecture and understanding different stages of plants, as result confirms the difference between drought and control plants based on architecture and number of leaves. Apart from this, we extracted features such as the  number of leaves detected and estimated by YOLO, area of convex hull formed by tips of leaves, horizontal spread of the plant, vertical spread of the plant and the ratio of the  the numbers of leaves to the area of convex hull to discriminate these genotypes. Figure \ref{fig:results_1_samples} shows detected leaf tips by YOLO and convex hull for control and drought conditions. Our model detects upto  90 percent of tips and leaves accurately. 

Figure \ref{fig:dendo} shows dendogram obtained using all the features i.e. the number of leaves in plant, area of convex hull of the plant, the number of leaves per plant per unit area convex hull of the plants, vertical and horizontal spread (spacing) of the plant Figure \ref{dendo1} and using only the number of leaves in a plants Figure, \ref{dendo2} based on controlled and drought conditions. Dendogram in Figure \ref{dendo1} is based on all parameters like the number of leaves,vertical and horizontal plant spread (considering top view of RGB image), the convex hull i.e. area under convex hull for each plants, and the number of leaves per unit area of convex hull of each plant. Results shows that genotypes  6 (DULAR), 7  (PMK-2) and 8 (SERATOES) fall in one group are very similar genotypes, same observation is confirmed in dendogram constructed based on the number of leaves, where 6, 7 and 8 are very in one group or similar genotypes. Genotype 9 (KALINGA-1) is distinct based on all parameters wheres considering the number of leaves genotype 4 (NAGINA 22) is distinct to others. These genotypes can be a basis for further genetic improvement programs.   

\section{Conclusions and Future Work}
Image-based phenotyping has an advantage over traditional phenotyping. Object detection techniques have potential in leaves counting tasks as we trained and demonstrated YOLO considering tip as an object in the current study. The number of leaves per convex hull of a plant is an inherent trait for image-based phenotyping. This can be the basis for monitoring the rate of leaf development in field crops like rice, wheat, maize in the early vegetative stages, and simulate it for different growth stages for different genotypes in a controlled environment. This can be used for further abiotic stress studies and phenotyping.
\subsubsection*{Acknowledgment}
Authors acknowledge scientists of Nanaji Deshmukh Plant Phenomics Center (NDPPC), IARI, New Delhi, India, and the ICAR-IARI for providing the high throughput image data to undertake this research work. The NDPPC is created and supported by National Agriculture Science Fund (NASF), ICAR, Grant No.NASF/Phen-6005/2016–17.
\bibliography{iclr2020_conference}
\bibliographystyle{iclr2020_conference}

\end{document}